\documentclass[11pt]{article} 
\usepackage{iclr2026_conference,times}  
\usepackage[hyphens]{url}  
\usepackage{graphicx} 
\usepackage{natbib}  
\usepackage{caption} 
\usepackage[margin=1in]{geometry}
\usepackage{amsmath}
\usepackage{amsfonts}
\usepackage{booktabs}
\usepackage{multirow}
\usepackage{makecell}
\usepackage{float}
\usepackage{tabularx}

\usepackage{subcaption}
\usepackage{xcolor}
\usepackage{hyperref}
\usepackage[nameinlink,capitalise]{cleveref}
\usepackage{url}
\usepackage{fancyvrb}
\usepackage{fvextra}
\usepackage{multicol}
\usepackage{enumitem}
\usepackage{xurl}
\usepackage[normalem]{ulem}

\usepackage[most]{tcolorbox}
\usepackage{listings}
\usepackage[table]{xcolor}

\newcommand{\op}[1]{\item \path{#1}}
\definecolor{crimson}{HTML}{DC143C}

\iclrfinalcopy

\definecolor{mtblue}{HTML}{C2410C}

\newcommand{\makeMusaTitle}{
\begin{center}

\vspace*{-6.3em}

\begin{minipage}{\textwidth}
    \raggedright
    \includegraphics[height=0.65cm]{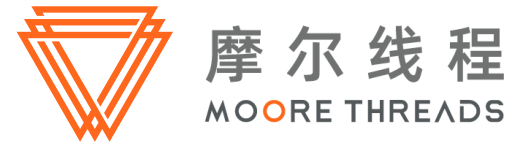}
\end{minipage}

\rule{\textwidth}{0.8pt}

{\Large\bfseries LEAP: Lean Environment-Feedback via Adaptive Pruning \\[0.4em] for Code RL in GPU Kernel Generation}

\vspace{1.0em}

{\large\bfseries
Tankun Li,\quad
Zhi Chen,\quad
Yaohua Tang
}

\vspace{0.1em}

{\normalsize
Moore Threads AI\\[0.25em]
\texttt{tangyaohua28@gmail.com}
}

\vspace{0.8em}

\end{center}
}

\begin{document}
\makeMusaTitle

\begin{abstract}
Post-training large language models (LLMs) via reinforcement learning (RL) has significantly advanced code generation capabilities. To bypass the heavy memory footprint of critic networks, current state-of-the-art frameworks leverage critic-free paradigms like Group Relative Policy Optimization (GRPO) tied to rule-based verification sandboxes. However, applying these frameworks to low-level systems programming, such as CUDA kernel generation—presents severe challenges: binary pass/fail rewards introduce severe signal sparsity, while multi-turn environmental feedback loops suffer from prohibitive compilation latencies and reward dilution across trajectories. 
In this work, we introduce \textbf{LEAP} (\textbf{L}ean \textbf{E}nvironment-Feedback via \textbf{A}daptive \textbf{P}runing), a scalable and computationally efficient multi-turn RL framework optimized for low-level hardware accelerator alignment. LEAP features \textbf{Difficulty-Conditioned Pruning (DCP)}, a dynamic gating mechanism that adaptively cuts off simple and overly catastrophic tasks from multi-turn expansion, focusing resource-heavy compilation and hardware exploration exclusively on high-value, complex tasks. To fully operationalize these paths without manual hyperparameter engineering, we propose a \textbf{Rank-Based Reward} formulation. By deriving scale-free relative advantages from pairwise tournament outcomes within the GRPO rollout group, our method inherently penalizes token inefficiency on simple prompts while maximizing learning gradients on challenging distributions. Empirical evaluations show that LEAP achieves superior first-turn proficiency and robust multi-turn debugging resilience while converging faster than unpruned multi-turn baselines, establishing a practical paradigm for low-level code RL.
\end{abstract}

\section{Introduction}
Large language models (LLMs) are central to modern software development, driving extensive research into LLM-based coding assistance. To enhance performance, recent work focuses on large-scale datasets for pre-training \cite{lozhkov2024starcoder} and supervised fine-tuning (SFT) \cite{wei2023magicoder}, evaluated against rigorous benchmarks \cite{liu2023your,jimenez2024swe}. Consequently, current state-of-the-art models deliver exceptional coding proficiency across diverse engineering tasks \cite{yang2025qwen3,zeng2026glm}.

Post-training models via Reinforcement Learning (RL) introduces memory bottlenecks. While Proximal Policy Optimization (PPO) \cite{schulman2017proximal} relies on a memory-intensive critic model, Group Relative Policy Optimization (GRPO) \cite{guo2025deepseek} eliminates this overhead by calculating relative advantages among multiple sampled rollouts per prompt. Because code correctness is deterministically verifiable via sandboxed unit testing, rule-based GRPO variants optimizing for binary rewards have become the mainstream approach to advancing code generation capabilities \cite{yu2026dapo,zheng2025group,cheng2026revisiting}. Despite its potential, rule-based RL suffers from reward sparsity, where binary feedback penalizes minor code errors as severely as catastrophic failures, prompting dense feedback mechanisms like token-level \cite{cheng2026reasoning} or step-level \cite{hou2025treerl} rewards that remain difficult to scale. Alternatively, leveraging stronger LLMs as judges offers softer, rubric-based grading \cite{gunjal2025rubrics,viswanathan2026checklists}, though this approach incurs significant computational costs and relies heavily on prompt engineering.

To establish a dense learning signal without external judging overhead, we turn to an alternative paradigm: multi-turn training with environmental feedback. Instead of merely penalizing failed samples, a multi-turn formulation leverages these feedback logs to provide richer iterative signals, closely mirroring how human engineers debug low-level systems code. While recent frameworks have explored multi-turn training on simplified academic benchmarks \cite{ekbote2025murphy,jain2025multi,gehring2024rlef}, two major issues hinder their practical deployment:

\noindent\hspace*{1.5em}\textbf{Reward Dilution Across Trajectories:} Typical multi-turn training evaluates the reward based solely on the final outcome and propagates it across the entire trajectory without differentiating per-turn performance. Consequently, a rollout requiring multiple debugging iterations receives the same reward as a rollout that succeeds on the first attempt. This biases the model toward the final result, greatly degrading its early-turn pass rate.

\noindent\hspace*{1.5em}\textbf{Prohibitive Compilation and Hardware Latency:} Multi-turn training is inherently time-consuming. Prior works typically formalize multi-turn generation as a tree-search over simple, interpreted languages (e.g., Python) that are incredibly fast to verify. In real-life production tasks such as from-scratch CUDA kernel generation, however, the overhead of sequential compilation, host-to-device memory allocation, and kernel execution introduces massive latency. Running dense, unpruned multi-turn rollouts under these conditions drastically slows down training iterations and makes large-scale code RL practically unviable.

In this work, we introduce \textbf{LEAP} (\textbf{L}ean \textbf{E}nvironment-Feedback via \textbf{A}daptive \textbf{P}runing), a scalable and efficient multi-turn RL framework designed specifically to address these hardware-level bottlenecks within a rule-based GRPO paradigm. Our framework is grounded in a simple intuition: harder coding tasks require richer learning signals, whereas simple tasks already provide sufficient feedback. We dynamically classify kernel difficulty based on the aggregate rollout pass rate. Simple problems, which exhibit high pass rates, naturally derive ample optimization signals from successful rollouts. Conversely, difficult problems suffer from a lack of successful trajectories. Leveraging this insight, LEAP employs a branching mechanism termed \textbf{Difficulty-Conditioned Pruning (DCP)}. DCP adaptively prunes simple problems from multi-turn expansion, bypassing redundant compilation cycles and focusing resource-heavy hardware exploration exclusively on complex tasks.

This adaptive pruning strategy yields three key advantages:

\noindent\hspace*{1.5em}\textbf{Computational Efficiency:} By restricting multi-turn branching via DCP, LEAP drastically lowers training time, saving substantial compute resources and GPU cycles otherwise wasted on redundant compilation and sandbox execution of trivial tasks.

\noindent\hspace*{1.5em}\textbf{Balanced Optimization:} The model learns to optimize early-turn accuracy on simpler kernels while gaining dense, multi-turn error-correction signals on complex ones, thereby preserving early-turn proficiency without sacrificing late-turn debugging capabilities.

\noindent\hspace*{1.5em}\textbf{Performance Retention:} We empirically demonstrate that our pruning strategy maintains—and in some cases surpasses—the performance of dense, unpruned multi-turn baseline frameworks.

To fully operationalize these efficiency gains, LEAP introduces a non-parametric, rank-based advantage formulation that generalizes across multi-turn trajectories. Rather than manually tuning heuristic, scalar-based ``magic numbers" for intermediary turns, our approach maps rollout outcomes directly to an ordinal hierarchy based on iteration efficiency. By deriving relative advantages strictly from pairwise win-loss distributions within the GRPO group, the resulting optimization signal becomes dynamically responsive to local prompt difficulty. For simpler prompts dominated by rapid success, the mathematical baseline shifts upward, penalizing excessive turns and driving the policy toward zero-shot efficiency. Conversely, for complex prompts characterized by low base pass rates, the framework naturally scales to reward incremental debugging breakthroughs, maximizing the gradient signal on hard samples. This formulation effectively mitigates reward dilution, enforces a strict temporal penalty that naturally minimizes inference tokens, and completely bypasses the instability of manual reward engineering.

\begin{figure}[htbp]
    \centering
    \includegraphics[width=0.8\linewidth]{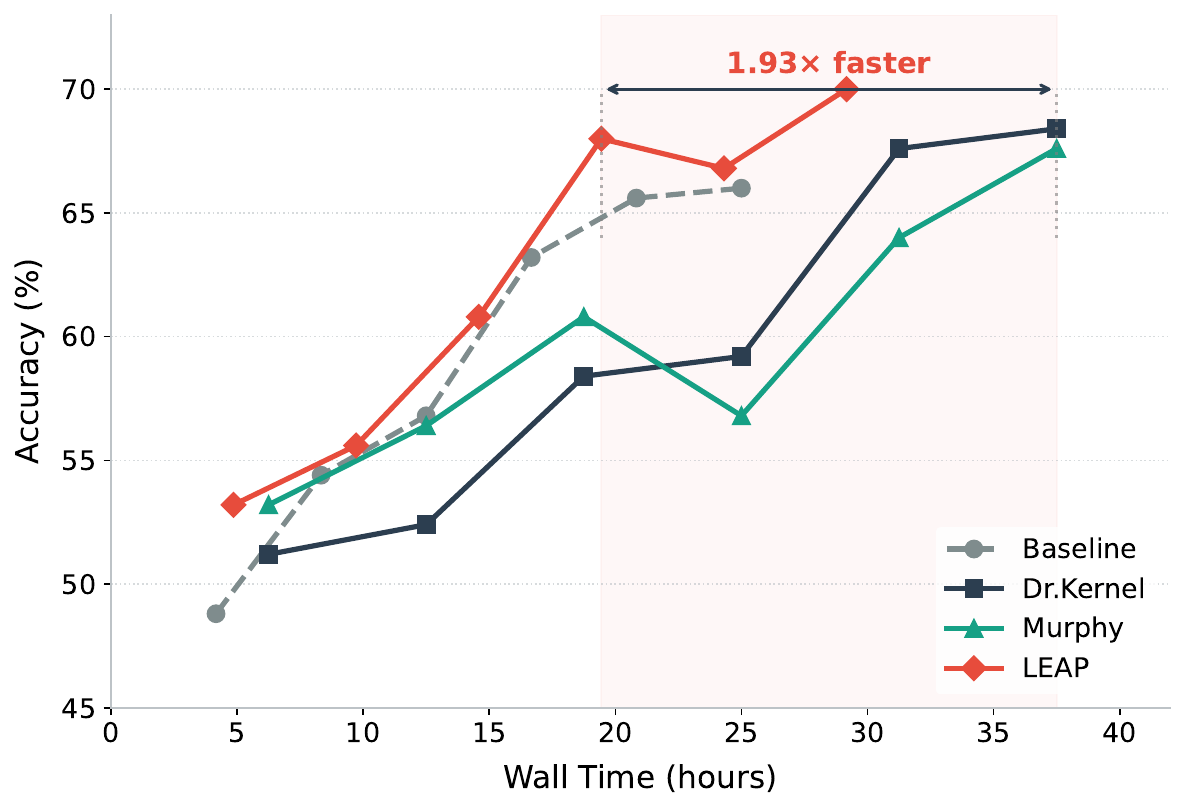}
    \caption{Test accuracy versus wall-clock training time. LEAP reaches the same accuracy roughly 1.93× faster than other works, and outperforms the Baseline at every wall-time budget.}
    \label{fig:intro}
\end{figure}

As illustrated in Fig.~\ref{fig:intro}, our method converges to similar performance 1.93$\times$ faster than existing approaches and outperform them in the end. In summary, our key contributions are:
\begin{itemize}
\item We propose \textbf{LEAP}, an efficient and scalable multi-turn RL framework optimized for low-level GPU kernel generation that drastically reduces the compilation and execution overhead typical of vanilla multi-turn RL methods.
\item We demonstrate through extensive experiments that our \textbf{Difficulty-Conditioned Pruning (DCP)} strategy does not degrade model performance; instead, it encourages the model to extract richer signals from hard samples while enhancing first-turn accuracy on simpler ones.
\item We introduce a scale-free, rank-based rewarding strategy within the GRPO paradigm that replaces rigid scalar rewards with relative pairwise advantages. This formulation dynamically enforces a temporal efficiency bias—incentivizing faster solutions on simple tasks while maximizing learning signals on complex, low-pass-rate benchmarks.
\item Our empirical results confirm that the LEAP framework achieves superior performance in both model capability and training efficiency for from-scratch CUDA kernel generation task and general coding task compared to existing baselines.
\end{itemize}

\section{Related Work}

\subsection{Code Reinforcement Learning for LLMs}
Post-training large language models via Reinforcement Learning (RL) has significantly advanced complex reasoning. Early paradigms relied on Proximal Policy Optimization (PPO) \cite{schulman2017proximal} using heavy value networks or Process Reward Models (PRMs) to supply continuous feedback \cite{yue2504vapo, cui2025process}, though at extreme computational and memory overhead. To eliminate critic-related bottlenecks, Group Relative Policy Optimization (GRPO) \cite{guo2025deepseek} estimates relative advantages across localized rollouts via deterministic, verifiable rewards. Subsequent extensions like Dr.GRPO \cite{liu2025understanding}, DAPO \cite{yu2026dapo}, and related industry systems \cite{xiao2026mimo, team2025kimi} focus on mitigating GRPO's inherent reward sparsity and stabilizing group variance, while intermediate frameworks incorporate Monte Carlo tree search \cite{hou2025treerl, yang2025treerpo} or LLM-as-a-judge rubrics \cite{gunjal2025rubrics, huang2025reinforcementlearningrubricanchors, viswanathan2026checklists, deepseek2026api}—yet these incur prohibitive latency and API costs.

Crucially, existing methodologies remain ill-equipped for multi-turn RL over interactive horizons. Current multi-turn approaches like $\mu$Code \cite{jain2025multi}, RLEF \cite{gehring2024rlef}, and REVEAL \cite{jin2025reveal} still depend on costly verifier or critic networks, while critic-free alternatives such as MURPHY \cite{ekbote2025murphy} suffer from unpruned exploration spaces that fail in production. LEAP directly solves these multi-turn scalability bottlenecks by introducing dynamic task-conditioned pruning within a critic-free framework.

\subsection{From-Scratch CUDA Kernel Generation}
The vast majority of reinforcement learning frameworks for code synthesis rely on academic-centric training datasets like KodCode \cite{xu2025kodcode} and evaluate performance on high-level benchmarks such as HumanEval \cite{liu2023your} and MBPP \cite{austin2021program}, which focus on lightweight interpreted languages for which modern foundational models \cite{yang2025qwen3, zeng2026glm} are already heavily optimized. To bridge this evaluative gap, our work targets from-scratch CUDA kernel generation. While KernelBench \cite{ouyang2025kernelbench} evaluates model proficiency in synthesizing CUDA code, its evaluation protocol lacks structural rigor, allowing models to artificially inflate pass rates by bypassing custom logic to invoke pre-existing PyTorch operators or optimized CUDA library wrappers. To enforce true architectural reasoning, we restrict our evaluation strictly to from-scratch kernel generation, strictly prohibiting external native implementations. A few parallel works have begun exploring the intersection of LLMs and accelerator optimization, but each introduces distinct inefficiencies. CUDA-L1 \cite{li2025cuda} adopts a standard GRPO framework but tasks the model with a detailed text-based analysis prior to code generation, creating a verbose execution overhead. Dr.Kernel \cite{liu2026dr} focuses on synthesizing Triton kernels utilizing the REINFORCE-with-leave-one-out (RLOO) framework \cite{ahmadian2024back}, but penalizes failed compilation attempts with non-informative zero rewards and naively permits every rollout to enter the debugging phase, which severely exacerbates the sandbox compilation latency bottleneck. Meanwhile, CUDA Agent \cite{dai2026cuda} proposes an agentic framework centered on optimizing kernel speeds using a traditional PPO backend, exposing the model to high-overhead tool environments while relying on empirical stabilizing heuristics rather than structural algorithmic efficiency. Distinct from these approaches, LEAP bypasses the hardware and compilation bottlenecks that restrict existing frameworks by introducing an efficient, critic-free multi-turn RL architecture tailored for low-level systems alignment.

\section{LEAP}
\begin{figure*}[!t]
    \centering
    \includegraphics[width=1.0\linewidth]{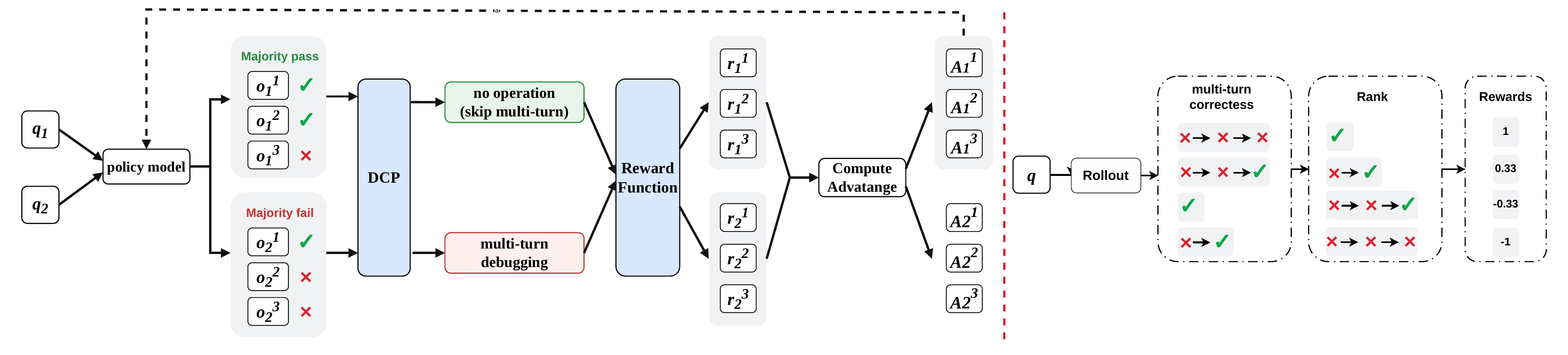}
    \caption{The overall structure of the LEAP (left) and rank-based reward (right). DCP refer to the difficulty conditioned pruning module. No-op stands for no operations (skip multi-turn debugging step). We skip the KL loss for simplicity.}
    \label{fig:overall}
\end{figure*}
Our framework is largely based on the standard GRPO baseline. The over all flow is illustrated in figure~\ref{fig:overall}. The differences between this work and standard GRPO is shown in the coloured box -- DCP and Reward Function. DCP act as a gating mechanism, controlling which groups should enter multi-turn debug step and the proposed reward function further balance the first-turn generation quality with iterative debugging capabilities. The details are explained in the following sections. 

\subsection{Difficulty-Conditioned Pruning (DCP)}
The Difficulty-Conditioned Pruning (DCP) module stems from an intuitive assumption regarding task complexity in Reinforcement Learning (RL): training distributions naturally comprise simpler tasks with high empirical pass rates and harder tasks with low pass rates. While simpler questions inherently enjoy dense learning signals derived from numerous successful optimization paths, more difficult questions suffer from sparse feedback, hindering gradual policy improvement. Under these conditions, uniformly applying multi-turn debugging across all tasks introduces two distinct negative effects: (1) for simpler tasks, errors in initial turns are insufficiently penalized, causing the model to over-rely on multi-turn corrections rather than converging on immediate correctness; and (2) it imposes a prohibitive computational burden on the training pipeline. To resolve these inefficiencies, we introduce DCP to selectively gate which questions transition into the multi-turn debugging environment based on task difficulty.

Let $q$ be a question sampled from the dataset distribution $P(Q)$, and let $\{o_i\}_{i=1}^G$ be a group of $G$ independent candidate responses sampled from the current policy $\pi_\theta(\cdot \mid q)$ within a group-based reinforcement learning framework. To quantify the empirical difficulty of $q$ at the current training step, DCP measures the cohort failure rate. Let $\mathbb{I}(\cdot)$ be an indicator function that returns $1$ if the response fails execution verification and $0$ if it passes. The total number of failed trajectories within a sampled group is defined as:
\begin{equation}
N_{\text{fail}}(q) = \sum_{i=1}^G \mathbb{I}(\text{Response } o_i \text{ failed})
\end{equation}

DCP introduces a conditional gating function $\mathcal{G}(q) \in \{0, 1\}$ that dynamically determines the computational path for the entire group:
\begin{equation}
\mathcal{G}(q) = \begin{cases} 
1, & \text{if } \tau_{\text{min}} \le N_{\text{fail}}(q) \le \tau_{\text{max}} \\
0, & \text{otherwise}
\end{cases}
\end{equation}
where $\tau_{\text{min}}$ and $\tau_{\text{max}}$ are strict hyperparameters defining the optimal learning zone for multi-turn optimization. When a question exhibits a high pass rate ($N_{\text{fail}}(q) < \tau_{\text{min}}$), it is classified as simple and bypasses the debugging loop ($\mathcal{G}(q) = 0$). This forces the model to learn from single-turn penalties, preventing it from developing a reliance on multi-turn corrections for easy tasks. Conversely, when a question exhibits a low pass rate ($N_{\text{fail}}(q) > \tau_{\text{max}}$), it is classified as overly difficult and the current model is hopeless to solve it. It similarly bypasses the debugging loop ($\mathcal{G}(q) = 0$) to protect compute resources from being wasted on unrecoverable trajectories. Only when the group failure count falls within the bounded threshold does $\mathcal{G}(q) = 1$, routing the group into the multi-turn debugging environment.

This high-level routing logic alters the standard GRPO objective by conditioning the optimization path on $\mathcal{G}(q)$. The objective function selectively scales the sequence length and computational layout based on this gate:
\begin{equation}
\begin{split}
L_{\text{DCP}}(\theta) = \, & \mathbb{E} \left[ q \sim P(Q), \{o_i\}_{i=1}^G \sim \pi_\theta \right] \\
& \left[ \frac{1}{G} \sum_{i=1}^G L_{\text{surr}}\left(\theta, o_i \;\middle|\; \mathcal{G}(q)\right) - \beta D_{\text{KL}}(\pi_\theta \parallel \pi_{\text{ref}}) \right]
\end{split}
\end{equation}
where the surrogate loss $L_{\text{surr}}$ evaluates a standard single-turn sequence when $\mathcal{G}(q) = 0$, and dynamically expands to evaluate a concatenated, multi-turn debugging sequence when $\mathcal{G}(q) = 1$. 

\subsection{Rank-Based Reward}
\label{sec:rank_based}

Empirically, we observed that varying the scalar reward assignments for multi-turn trajectories yields highly divergent optimization outcomes, requiring extensive and costly ablation studies to discover an optimal configuration. To fully operationalize the training paths established by the DCP module without introducing the optimization instability of hand-tuned heuristic values, LEAP introduces a non-parametric, rank-based reward framework coupled with GRPO. Instead of assigning continuous scalar ``magic numbers'' to different rollouts, our approach derives optimization gradients by first computing a baseline-free reward from the relative pairwise win-loss distribution within each rollout group, mapping outcomes directly to an execution efficiency hierarchy.

Depending on the gating state $\mathcal{G}(q)$, candidate responses enter an execution environment allowed up to $M$ maximum turns. We rank all the rollouts in the same group according to:
\begin{equation}
\mathcal{O}_{1\text{-st turn pass}} \succ \mathcal{O}_{2\text{-nd turn pass}} \succ \dots \succ \mathcal{O}_{\text{failure}}
\end{equation}
where $\succ$ denotes strict preference. That is, the rollouts use less turns to succeed are preferred over those use more turns. Specifically, we assign discrete rank values $r(o_i) \in \mathbb{N}$. A response that achieves successful verification on the first turn receives the highest rank $r = M$, a successful correction on the $m$-th turn receives $r = M - m + 1$, and a persistent execution or compilation failure at the end of the trajectory maps to the lowest rank $r = 0$.

For any candidate output $o_i$ within a group of size $G$, we calculate its baseline-free reward $R_i$ by performing a complete pairwise tournament against its $G-1$ local peers. We define the count of strictly worse peers $N_{\text{worse}}(o_i)$ and strictly better peers $N_{\text{better}}(o_i)$ as:
\begin{equation}
N_{\text{worse}}(o_i) = \sum_{j \neq i}^G \mathbb{I}\Big(r(o_i) > r(o_j)\Big), \quad N_{\text{better}}(o_i) = \sum_{j \neq i}^G \mathbb{I}\Big(r(o_i) < r(o_j)\Big)
\end{equation}

The pairwise tournament reward $R_i$ is computed as the net peer matchup score, normalized exclusively by the competing peer cohort size $G-1$:
\begin{equation}
R_i = \frac{N_{\text{worse}}(o_i) - N_{\text{better}}(o_i)}{G - 1}
\end{equation}

where the calculated reward $R_i$ is bounded strictly to the closed interval $[-1.0, 1.0]$. This normalization guarantees that an absolute winner within the rollout group receives a reward of exactly $+1.0$, while an absolute loser is anchored firmly to $-1.0$.

Finally, to optimize the policy, we compute the final advantage $A_i$ applied to the surrogate loss $\mathcal{L}_{\text{surr}}$ using the standard GRPO formulation.

The design of our rank-based framework deliberately eliminates reward hyperparameter tuning overhead. Compared to rigid, hand-tuned heuristic reward functions, our method offers a streamlined alternative that achieves two critical objectives: it inherently incentivizes execution efficiency by penalizing unnecessary reasoning turns, and it self-adaptively adjusts relative reward scaling based on the empirical difficulty of the problem context.

To systematically demonstrate the self-adaptive properties of the rank-based reward framework against static heuristic assignments, we analyze a restricted three-tier trajectory setting. Let the maximum number of turns be bounded to $M=2$, naturally segmenting the trajectory outcome space into three distinct tiers: first-turn success ($\mathcal{O}_{1}$), second-turn success ($\mathcal{O}_{2}$), and failure ($\mathcal{O}_{3}$). 

Let $N_1, N_2,$ and $N_3$ denote the frequency of responses landing in $\mathcal{O}_1, \mathcal{O}_2,$ and $\mathcal{O}_3$ respectively within a rollout group of size $G$, such that $G = N_1 + N_2 + N_3$. Under the proposed pairwise tournament framework, the baseline-free reward $R_i$ for an individual response $o_i$ in each respective tier is formulated as:
\begin{align}
    R(\mathcal{O}_1) &= \frac{N_2 + N_3}{G - 1} \label{eq:r1} \\
    R(\mathcal{O}_2) &= \frac{N_3 - N_1}{G - 1} \label{eq:r2} \\
    R(\mathcal{O}_3) &= \frac{-(N_1 + N_2)}{G - 1} \label{eq:r3}
\end{align}

The critical distinction between our non-parametric framework and standard hand-tuned rewards lies in the mathematical behavior of the intermediate tier, $R(\mathcal{O}_2)$. In a traditional reinforcement learning setup, intermediate successes are typically assigned a static scalar hyperparameter $c \in (0, 1)$ (e.g., $R_{\text{static}}(\mathcal{O}_2) = c$). This static assignment fundamentally assumes that the value of a multi-turn correction is independent of the problem's inherent difficulty. 

Conversely, our rank-based formulation in Equation \ref{eq:r2} reveals that $R(\mathcal{O}_2)$ is strictly a function of the divergence between the failure density $N_3$ and the optimal success density $N_1$. This yields a mathematically rigorous adaptive curriculum that scales dynamically with empirical prompt difficulty:

\paragraph{Regime 1: Low-Difficulty Contexts ($N_1 > N_3$).} 
When the problem is easy, the empirical distribution skews heavily toward first-turn successes. Consequently, the intermediate reward satisfies:
\begin{equation}
    \lim_{N_1 \gg N_3} R(\mathcal{O}_2) < 0
\end{equation}
In this regime, requiring a second turn indicates relative execution inefficiency. The rank-based framework automatically penalizes $\mathcal{O}_2$, driving the policy gradient toward optimal first-turn efficiency.

\paragraph{Regime 2: High-Difficulty Contexts ($N_3 > N_1$).}
When the problem is exceedingly difficult, the empirical distribution is dominated by compilation or execution failures. Under these conditions, the intermediate reward satisfies:
\begin{equation}
    \lim_{N_3 \gg N_1} R(\mathcal{O}_2) > 0
\end{equation}
Because first-turn successes are rare ($N_1 \to 0$), successfully correcting an error on the second turn represents a significant statistical achievement relative to the peer cohort. The rank-based reward dynamically scales toward $+1.0$, heavily rewarding the correction capability and preventing gradient collapse.

In summary, because the partial derivatives satisfy $\frac{\partial R(\mathcal{O}_2)}{\partial N_3} > 0$ and $\frac{\partial R(\mathcal{O}_2)}{\partial N_1} < 0$, the intermediate reward acts as a continuous, baseline-free interpolator on $[-1, 1]$. By mathematically coupling the reward of partial successes to the underlying outcome distribution of the peer cohort, LEAP wholly eliminates the optimization instability and hyperparameter search space intrinsic to static, hand-tuned heuristic values.

\subsection{Discounted Per-Turn Advantage Assignment}
Given a rollout-level advantage $A_i$ for rollout $i$ computed as above-mentioned, we assign token-level advantages according to the turn structure of the generated trajectory. Let the trajectory contain $T_i$ model-generated turns, indexed by $t \in \{0, 1, \ldots, T_i - 1\}$. Let $\mathcal{M}_{i,t}$ denote the set of response tokens in rollout $i$ that belong to generated turn $t$. Tokens outside all generated turns, such as padding, prompt tokens, or environment feedback tokens, are excluded from optimization and receive zero advantage.

For each rollout, we first choose a discount base according to the sign of its rollout-level advantage, where $\gamma \in (0,1]$:
\[
\beta_i =
\begin{cases}
\gamma, & A_i \ge 0, \\[4pt]
1, & A_i < 0.
\end{cases}
\]
The advantage assigned to every token in turn $t$ is then the rollout-level advantage multiplied by the turn discount:
\[
\hat{A}_{i,k} = \beta_i^{T_i - t - 1} A_i, \qquad k \in \mathcal{M}_{i,t}.
\]
All tokens in the same generated turn therefore share the same advantage. The last generated turn has exponent zero and receives the original rollout-level advantage $A_i$. Earlier turns receive progressively smaller positive advantages when $A_i \ge 0$. When $A_i < 0$, the discount base is $1$, so every generated turn receives the same negative advantage $A_i$.

This assignment keeps the rollout-level preference signal unchanged while distributing it across turns. Positive trajectories place more credit on later turns, which are often closer to the final successful repair in a debugging trajectory. Negative trajectories apply the same penalty to all generated turns, discouraging the full sequence of actions that led to the unfavorable outcome.

\section{Experiments}
\subsection{Implementation Details}
We conduct experiments based on VERL \cite{sheng2024hybridflow}. The experiments, including ablation studies, are mainly performed on CUDA generation task. For the sake of reproducibility, we also report the performance in general coding task using public data. For from-scratch CUDA kernel generation task, we perform cold-start training on pytorch-to-CUDA dataset \cite{cheng2026musacoder} to establish a foundational knowledge base in CUDA programming. Subsequently, we apply Reinforcement Learning (RL) training using the CUDA-Agent dataset \cite{dai2026cuda}, validating the model within a custom-built sandbox that provides detailed execution tracebacks. For benchmarking, we evaluate our approach on KernelBench \cite{ouyang2025kernelbench}. For general coding task, we follow KodCode implementation \cite{xu2025kodcode} and perform RL with Qwen2.5-7B on KodCode dataset. The performances are reported with KodCode test set and LiveCodeBench \cite{jain2025livecodebench}.

When comparing with existing works, we choose standard GRPO as a baseline and compare our work with MURPHY \cite{ekbote2025murphy} as well as Dr.Kernel \cite{liu2026dr}, two GRPO-based multi-turn training frameworks. For MURPHY implementation, it is infeasible to perform a complete tree-search for multi-turn training in practice due to massive sandbox verification cost, we restrict the leaf node branching to one. 

During RL training, the hyperparameters are configured with a learning rate of $1\times10^{-6}$, a weight decay of $0.1$, a batch size of $32$, and $8$ rollouts per prompt. For the generation settings, the sampling temperature, top\_k, and top\_p are set to $1$, $-1$, and $0.95$ during rollout, and $0.7$, $20$, and $0.95$ during validation, respectively. Because our training framework mixes single-turn and multi-turn data, the variance in training sequence lengths can be quite high. To prevent lengthy multi-turn trajectories from disproportionately dominating each gradient update, we adopt the sample-level loss objective proposed in DAPO \cite{yu2026dapo}. For a fair comparison, all baseline experiments are conducted under this identical setting.

We perform all the training on a server with 8 Nvidia B200 GPUs and the verification sandbox for CUDA kernel is deployed on two server with 8 Nvidia A100 GPUs. For Kodcode-style training, we follow the official implementation carefully.

\subsection{Main Results}
\subsubsection{KernelBench}

\begin{table}[t]
\centering
\caption{Model performance across methods and difficulty levels on KernelBench}
\label{tab:model_comparison}
\resizebox{0.8\linewidth}{!}{
\begin{tabular}{lc cccc}
\toprule
\textbf{Level} & \textbf{Turn} & \textbf{Baseline} & \textbf{Murphy} & \textbf{Dr.Kernel} & \textbf{LEAP} \\
\midrule
\multirow{3}{*}{Level 1 (\%)} & 1 & 80 & 79 & \textbf{84} & \textbf{84} \\
                         & 2 & 91 & 92 & \textbf{95} & 93 \\
                         & 3 & 91 & 93 & \textbf{96} & 95 \\
\midrule
\multirow{3}{*}{Level 2 (\%)} & 1 & 76 & \textbf{78} & 75 & 76 \\
                         & 2 & 85 & 82 & \textbf{88} & \textbf{88} \\
                         & 3 & 85 & 83 & \textbf{89} & \textbf{89} \\
\midrule
\multirow{3}{*}{Level 3 (\%)} & 1 & 18 & 24 & 24 & \textbf{30} \\
                         & 2 & 24 & \textbf{34} & 32 & \textbf{34} \\
                         & 3 & 24 & 34 & 32 & \textbf{36} \\
\midrule
\multirow{3}{*}{Overall (\%)} & 1 & 66 & 67.6 & 68.4 & \textbf{70} \\
                         & 2 & 75.2 & 76.4 & \textbf{79.6} & 79.2 \\
                         & 3 & 75.2 & 77.2 & 80.4 & \textbf{80.8} \\
\bottomrule
\addlinespace
\end{tabular}
}
\end{table}


\begin{table}[t]
    \centering
    \caption{Training cost and turn-efficiency of different methods.}
    \label{tab:method_performance_rotated}
    \begin{tabularx}{0.8\columnwidth}{l *{4}{>{\centering\arraybackslash}X}} 
        \toprule
        \textbf{Method} & Baseline & Murphy & Dr.Kernel & LEAP \\
        \midrule
        \textbf{Step Time (s)} & $\sim$600 & $\sim$950 & $\sim$950 & $\sim$700 \\
        \textbf{Turn Per Pass} & 1.90 & 1.93 & 1.84 & 1.77 \\
        \bottomrule
    \end{tabularx}
\end{table}

The empirical evaluation of LEAP against existing baselines is summarized in Table~\ref{tab:model_comparison}. Models are evaluated across a maximum of three turns, with cumulative accuracy reported at each iteration. Notably, LEAP achieves the highest first-turn accuracy among all tested frameworks. Both LEAP and Dr.Kernel consistently outperform the remaining baselines. However, compared to Dr.Kernel, LEAP yields superior results in the first turn while maintaining highly competitive performance in subsequent turns, demonstrating its ability to preserve zero-shot proficiency without sacrificing multi-turn error correction.

Beyond model capabilities, LEAP delivers substantial computational savings. As detailed in Table~\ref{tab:method_performance_rotated}, our framework requires less training time than competing methods. Using first-turn accuracy as a metric to measure model convergence, we plot accuracy versus wall-clock time in Fig.~\ref{fig:intro}. Notably, LEAP demonstrates wall-clock efficiency, reaching comparable convergence $1.93\times$ faster than alternative methods. To report ``How many total turns did this run spend for each successful sample it got?'', the turn-efficiency is also reported in Table~\ref{tab:method_performance_rotated}, which is computed by (total turns used by all samples) / (passed samples). LEAP demonstrate the best turn-efficiency among all the peers, using less turns on average to solve the problems.

Taken together, these results validate that LEAP effectively balances optimization across varying problem difficulties—safeguarding first-turn performance, retaining robust multi-turn debugging capabilities, and drastically accelerating training throughput.

\subsubsection{KodCode}
\begin{table}[t]
\centering
\caption{Performance comparison across different methods on KodCode test set}
\label{tab:kod_performance_comparison}
\begin{tabular}{cccc}
\toprule
\textbf{Turn} & \textbf{Murphy} & \textbf{Dr.Kernel} & \textbf{LEAP} \\
\midrule
1 & 88.5 & 88.3 & \textbf{90.2} \\
2 & 94.5 & 94.1 & \textbf{94.7} \\
3 & \textbf{96.7} & 96.5 & \textbf{96.7} \\
\bottomrule
\end{tabular}
\end{table}

\begin{figure}[t]
    \centering
    \includegraphics[width=0.8\linewidth]{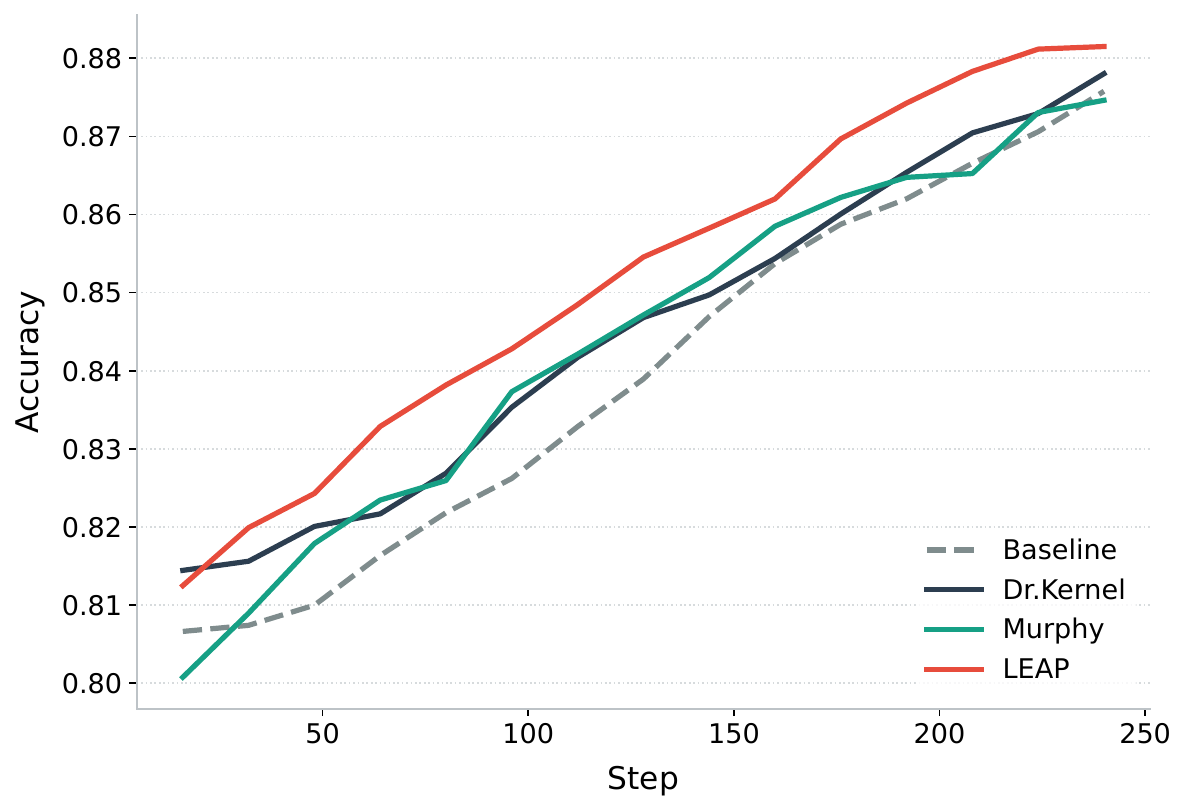}
    \caption{The first turn accuracy v.s. training steps on KodCode test set. The values are smoothed with factor 0.8.}
    \label{fig:kodcode}
\end{figure}

We also conducted evaluation experiments on the KodCode dataset \cite{xu2025kodcode}, with the corresponding results presented in Table~\ref{tab:kod_performance_comparison}. Because KodCode is a relatively straightforward, Python-centric dataset, performance across subsequent iterations (Turns 2 and 3) remains highly comparable. Notably, the primary advantage of LEAP lies in its superior first-turn accuracy, where it achieves the top performance. To further illustrate this advantage, we plot the first-turn accuracy against the training steps in Figure~\ref{fig:kodcode}. As the curves demonstrate, LEAP consistently outperforms competing methods throughout the entire training duration.

\subsubsection{LiveCodeBench}
\begin{table}[t]
\centering
\caption{Performance comparison across different methods on LiveCodeBench}
\label{tab:lcb}
\resizebox{0.8\columnwidth}{!}{%
\begin{tabular}{lcccc}
\toprule
\textbf{Method} & \textbf{Baseline} & \textbf{Murphy} & \textbf{Drkernel} & \textbf{LEAP} \\
\midrule
Easy Pass@1   & 0.609 & 0.614 & 0.609 & \textbf{0.619} \\
Medium Pass@1 & \textbf{0.088} & 0.085 & 0.085 & 0.085 \\
Hard Pass@1   & 0.007 & 0.010 & 0 & \textbf{0.017} \\
Pass@1        & 0.174 & 0.181 & 0.175 & \textbf{0.185} \\
\midrule
Easy Pass@5   & 0.744 & 0.767 & 0.767 & \textbf{0.814} \\
Medium Pass@5 & 0.173 & \textbf{0.192} & 0.153 & 0.173 \\
Hard Pass@5   & 0.038 & \textbf{0.05} & 0.041 & \textbf{0.05} \\
Pass@5        & 0.251 & 0.269 & 0.234 & \textbf{0.274} \\
\bottomrule
\end{tabular}%
}
\end{table}

After training on KodCode dataset, we follow the previous work to evaluate the model on LiveCodeBench in Table~\ref{tab:lcb}. LEAP demonstrate the best overall performance among peers, which suggest LEAP's potential to be applied to general coding tasks.

\subsection{Study on Branch Pruning}

\begin{table}
\centering
\caption{Impact of pruning range on performance}
\label{tab:pruning_groups}
\begin{tabularx}{0.8\columnwidth}{>{\raggedright\arraybackslash}X c ccc}
\toprule
\textbf{Metric} & \textbf{Baseline} & $\mathbf{\le50\%}$ & $\mathbf{\ge50\%}$ & \textbf{Full} \\
\midrule
Acc@3-turns       & 75.2 & 77.6 & 80.8 & 80.4 \\
\addlinespace
Activated groups     & -- & 57.1\% & 43.9\% & 42.3\% \\
\addlinespace
Per-rollout Recovery & -- & 43.3\% & 20.5\% & 26.4\% \\
\addlinespace
Per-group Recovery   & -- & 73.9\% & 61.8\% & 64.5\% \\
\bottomrule
\end{tabularx}
\end{table}

In this section, we analyze the impact of different pruning ranges within the DCP module, with the experimental results summarized in Table~\ref{tab:pruning_groups}. The ``Pruning Fail Range'' specifies the pass-rate thresholds that dictate whether a rollout group enters the multi-turn debugging phase. Specifically, $\le$50\% targets less challenging problems (where the initial pass-rate is $\ge$50\%), while $\ge$50\% isolates more difficult problems (where the initial pass-rate is $\le$50\%). The ``Full'' setting denotes a baseline configuration where all problems bypass pruning and proceed directly to the multi-turn phase.

To evaluate these dynamics, we track several key metrics during training. ``Activated groups'' represents the percentage of total rollout groups that trigger the multi-turn debugging phase. ``Per-rollout Recovery Rate'' measures the proportion of individual multi-turn trajectories that successfully recover from an initial execution failure, whereas ``Per-group Recovery Rate'' denotes the ratio of groups that yield at least one successfully debugged trajectory.

Our findings reveal that deploying the multi-turn debugging phase exclusively on harder problems ($\ge$50\%) yields superior final accuracy compared to focusing on easier problems ($\le$50\%), despite the fact that easier problems inherently exhibit higher baseline recovery rates. This strongly validates our core hypothesis: for code generation tasks, simpler problems already benefit from dense learning signals provided by an abundance of successful initial trajectories. Conversely, more difficult problems suffer from sparse signals. By selectively routing only the harder problems into the multi-turn debugging pipeline, the model encounters a significantly more effective learning gradient, maximizing performance gains while avoiding the prohibitive training costs associated with the ``Full'' multi-turn configuration.

\subsection{Study on Reward functions}

\begin{table}[t]
    \centering
    \caption{Rank-based reward function comparison}
    \label{tab:rank_based}
    \begin{tabular}{lccc}
        \toprule
        \textbf{Setting} & \textbf{Turn} & \textbf{GRPO-MT} & \textbf{Rank-based} \\
        \midrule
        \multirow{3}{*}{Level 1 (\%)} & 1 & 83 & \textbf{84} \\
                                 & 2 & 91 & \textbf{93} \\
                                 & 3 & 91 & \textbf{95} \\
        \midrule
        \multirow{3}{*}{Level 2 (\%)} & 1 & \textbf{79} & 76 \\
                                 & 2 & 86 & \textbf{88} \\
                                 & 3 & \textbf{89} & \textbf{89} \\
        \midrule
        \multirow{3}{*}{Level 3 (\%)} & 1 & 26 & \textbf{30} \\
                                 & 2 & 32 & \textbf{34} \\
                                 & 3 & \textbf{36} & \textbf{36} \\
        \midrule
        \multirow{3}{*}{Overall (\%)} & 1 & \textbf{70.0} & \textbf{70.0} \\
                                 & 2 & 77.2 & \textbf{79.2} \\
                                 & 3 & 79.2 & \textbf{80.8} \\
        \bottomrule
    \end{tabular}
\end{table}

In this section, we isolate and analyze the specific impact of our non-parametric, rank-based advantage formulation. As formalized in Section~\ref{sec:rank_based} and further analyzed in Supplement, this rewarding paradigm dynamically scales advantages based on local prompt difficulty, naturally regularizing the policy to prioritize minimal-turn solutions. In the table, GRPO-MT stands for regular GRPO multi-turn training with standard advantages computation.

The empirical evidence for this behavior is presented in Table~\ref{tab:rank_based}. Compared to standard GRPO loss, which relies on rigid, static scalar rewards, our rank-based approach yields increases in each turn accuracy. This gap demonstrates that the dynamically shifting baseline successfully penalize verbose or inefficient trajectories on simpler tasks while preserve the gradient signal required for incremental debugging on complex ones. Specifically, in the early training where the model is under-performing, the multi-turn trajectories provide rich signals and it is rewarded heavily, whereas in the later stages where the model is performing well, the redundant multi-turn trajectories will be punished for those easier tasks.

Consequently, these findings confirm that transitioning from heuristic scalar rewards to relative pairwise advantages is essential for aligning multi-turn reinforcement learning with inference-token efficiency, effectively mitigating the trade-off between zero-shot proficiency and multi-turn resilience.

\subsection{Training Observation}
\begin{figure}
    \centering
    \includegraphics[width=0.5\linewidth]{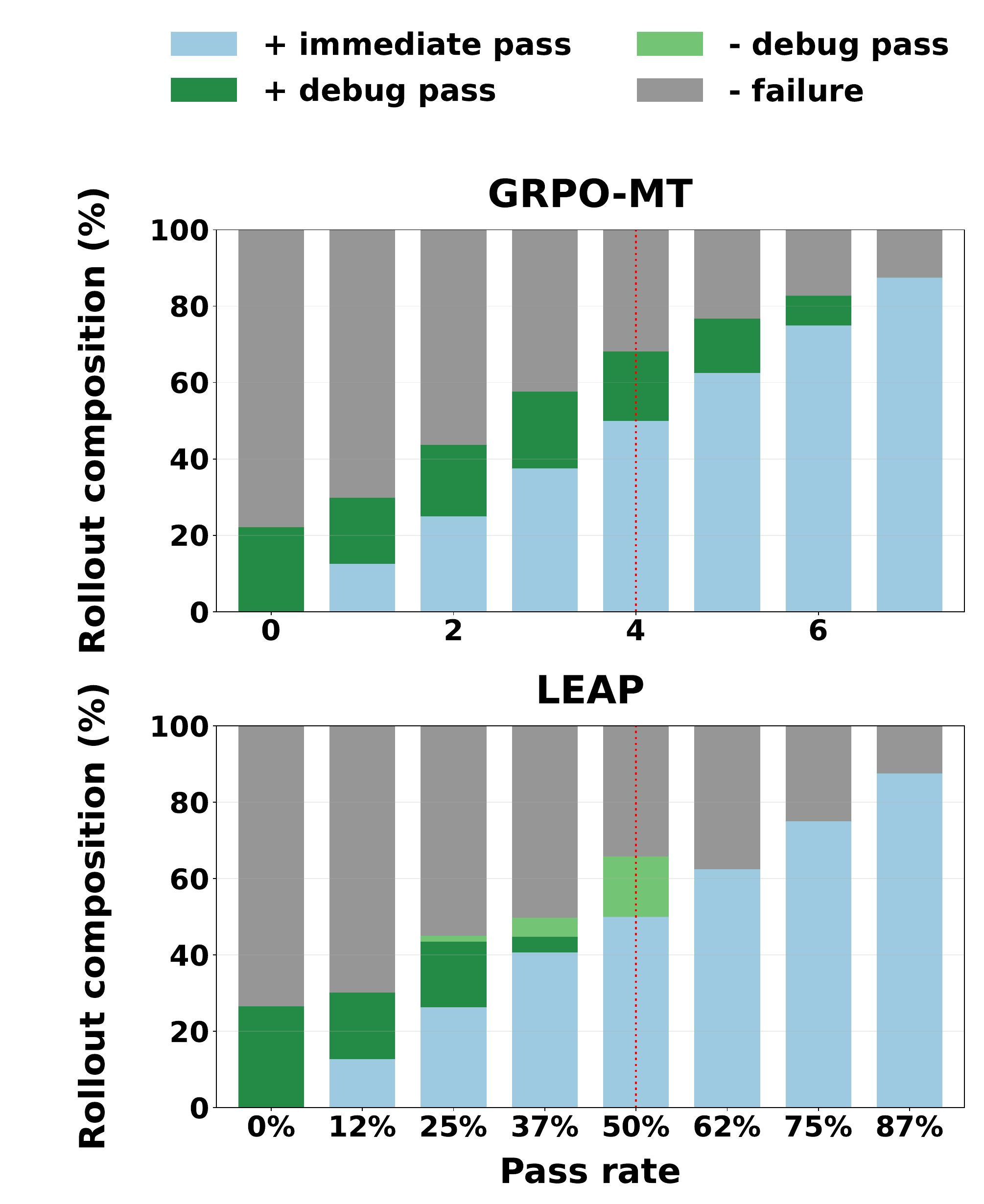}
    \caption{The rollout composition during training of LEAP and GRPO-MT.}
    \label{fig:rollout}
\end{figure}

Figure~\ref{fig:rollout} presents the rollout-state composition of effective training samples under GRPO multi-turn and LEAP. Rollouts are grouped by problem instance and training step, and their advantages are computed from the corresponding reward definitions. The rollouts are bucketed by observed pass rate and categorized by outcome type: immediate pass, debug-assisted pass, or failure, with the sign of the advantage indicated in the legend. The y-axis reports the percentage composition of nonzero-advantage rollouts within each pass-rate bucket. The x-axis denotes the observed pass rate of the rollout group, used as a proxy for sample difficulty: lower pass rates correspond to harder samples, while higher pass rates correspond to easier samples. As shown in the figure, easier problems already have higher first-turn pass rates and therefore receive dense learning signals from immediate successes, while the additional signal contributed by multi-turn recoveries is relatively small. This causes GRPO-MT to place a disproportionate amount of effective training signal on easier problems. In contrast, LEAP uses the DCP mechanism to gate easier problems from entering additional debugging turns, leading to a more balanced distribution of learning signal across difficulty levels. This more targeted allocation of training signal helps LEAP converge faster and reach a better optimum.

\section{Conclusion}
In this work, we presented LEAP, an efficient, critic-free multi-turn reinforcement learning framework designed specifically to overcome the computational and architectural bottlenecks of from-scratch CUDA kernel generation. By introducing Difficulty-Conditioned Pruning (DCP), we successfully converted the traditionally expensive multi-turn tree-search into an adaptive execution path, isolating resource-intensive compilation and hardware sandboxing onto tasks where dense signals are critically required. Furthermore, our non-parametric, rank-based reward framework eliminates the unstable search space of hand-tuned scalar rewards. Through a localized pairwise tournament scheme within the GRPO paradigm, the optimization objective self-adaptively shifts its baseline to penalize sub-optimal token efficiency on simple tasks while amplifying sparse learning gradients on complex problems. Our empirical results on low-level systems benchmarks demonstrate that LEAP significantly mitigates the traditional trade-offs of multi-turn code RL, preserving exceptional first-turn proficiency while retaining robust error-correction capabilities. By drastically reducing runtime latencies without sacrificing alignment quality, LEAP establishes a highly scalable and viable paradigm for post-training LLMs in specialized, computationally intensive hardware optimization domains.

\bibliographystyle{plainnat}
\bibliography{sample}

\begin{thebibliography}{37}
\providecommand{\natexlab}[1]{#1}
\providecommand{\url}[1]{\texttt{#1}}
\expandafter\ifx\csname urlstyle\endcsname\relax
  \providecommand{\doi}[1]{doi: #1}\else
  \providecommand{\doi}{doi: \begingroup \urlstyle{rm}\Url}\fi

\bibitem[Ahmadian et~al.(2024)Ahmadian, Cremer, Gall{\'e}, Fadaee, Kreutzer, Pietquin, {\"U}st{\"u}n, and Hooker]{ahmadian2024back}
Arash Ahmadian, Chris Cremer, Matthias Gall{\'e}, Marzieh Fadaee, Julia Kreutzer, Olivier Pietquin, Ahmet {\"U}st{\"u}n, and Sara Hooker.
\newblock Back to basics: Revisiting reinforce-style optimization for learning from human feedback in llms.
\newblock In \emph{Proceedings of the 62nd Annual Meeting of the Association for Computational Linguistics (Volume 1: Long Papers)}, pages 12248--12267, 2024.

\bibitem[Austin et~al.(2021)Austin, Odena, Nye, Bosma, Michalewski, Dohan, Jiang, Cai, Terry, Le, et~al.]{austin2021program}
Jacob Austin, Augustus Odena, Maxwell Nye, Maarten Bosma, Henryk Michalewski, David Dohan, Ellen Jiang, Carrie Cai, Michael Terry, Quoc Le, et~al.
\newblock Program synthesis with large language models.
\newblock \emph{arXiv preprint arXiv:2108.07732}, 2021.

\bibitem[Cheng et~al.(2026{\natexlab{a}})Cheng, Huang, Zhu, Dai, Zhao, Zhang, and Wei]{cheng2026reasoning}
Daixuan Cheng, Shaohan Huang, Xuekai Zhu, Bo~Dai, Xin Zhao, Zhenliang Zhang, and Furu Wei.
\newblock Reasoning with exploration: An entropy perspective.
\newblock In \emph{Proceedings of the AAAI Conference on Artificial Intelligence}, volume~40, pages 30377--30385, 2026{\natexlab{a}}.

\bibitem[Cheng et~al.(2026{\natexlab{b}})Cheng, Hao, Liu, Zhou, Xie, Yao, Bian, Dey, Zhuang, Zha, et~al.]{cheng2026revisiting}
Jorge~Zhoujun Cheng, Shibo Hao, Tianyang Liu, Fan Zhou, Yutao Xie, Feng Yao, Yuexin Bian, Nilabjo Dey, Yonghao Zhuang, Yuheng Zha, et~al.
\newblock Revisiting reinforcement learning for llm reasoning from a cross-domain perspective.
\newblock \emph{Advances in Neural Information Processing Systems}, 38, 2026{\natexlab{b}}.

\bibitem[Cheng et~al.(2026{\natexlab{c}})Cheng, Lu, Liao, Li, Zhang, Yang, Lv, Wang, Chen, and Tang]{cheng2026musacoder}
Kun Cheng, Songshuo Lu, Sicong Liao, Tankun Li, Yafei Zhang, Dong Yang, Qiheng Lv, Hua Wang, Zhi Chen, and Yaohua Tang.
\newblock Musacoder: Native gpu kernel generation with full-stack training on moore threads gpu.
\newblock \emph{arXiv preprint arXiv:2606.04847}, 2026{\natexlab{c}}.

\bibitem[Cui et~al.(2025)Cui, Yuan, Wang, Wang, Zhang, Chen, Li, He, Fan, Yu, et~al.]{cui2025process}
Ganqu Cui, Lifan Yuan, Zefan Wang, Hanbin Wang, Yuchen Zhang, Jiacheng Chen, Wendi Li, Bingxiang He, Yuchen Fan, Tianyu Yu, et~al.
\newblock Process reinforcement through implicit rewards.
\newblock \emph{arXiv preprint arXiv:2502.01456}, 2025.

\bibitem[Dai et~al.(2026)Dai, Wu, Yu, Gao, Li, Jiang, Lou, Song, Yu, Chen, et~al.]{dai2026cuda}
Weinan Dai, Hanlin Wu, Qiying Yu, Huan-ang Gao, Jiahao Li, Chengquan Jiang, Weiqiang Lou, Yufan Song, Hongli Yu, Jiaze Chen, et~al.
\newblock Cuda agent: Large-scale agentic rl for high-performance cuda kernel generation.
\newblock \emph{arXiv preprint arXiv:2602.24286}, 2026.

\bibitem[{DeepSeek AI}(2026)]{deepseek2026api}
{DeepSeek AI}.
\newblock {DeepSeek API News Update}.
\newblock \url{https://api-docs.deepseek.com/news/news260424}, 2026.
\newblock Accessed: 2026-05-26.

\bibitem[Ekbote et~al.(2025)Ekbote, Lingam, Tehrani, Huan, Sanghavi, Deoras, and Soatto]{ekbote2025murphy}
Chanakya Ekbote, Vijay Lingam, Behrooz~Omidvar Tehrani, Jun Huan, Sujay Sanghavi, Anoop Deoras, and Stefano Soatto.
\newblock Murphy: Reflective multi-turn reinforcement learning for self-correcting code generation in large language.
\newblock In \emph{First Workshop on Foundations of Reasoning in Language Models}, 2025.
\newblock URL \url{https://openreview.net/forum?id=x0Ir7cWEiA}.

\bibitem[Gehring et~al.(2024)Gehring, Zheng, Copet, Mella, Carbonneaux, Cohen, and Synnaeve]{gehring2024rlef}
Jonas Gehring, Kunhao Zheng, Jade Copet, Vegard Mella, Quentin Carbonneaux, Taco Cohen, and Gabriel Synnaeve.
\newblock Rlef: Grounding code llms in execution feedback with reinforcement learning.
\newblock \emph{arXiv preprint arXiv:2410.02089}, 2024.

\bibitem[Gunjal et~al.(2025)Gunjal, Wang, Lau, Nath, He, Liu, and Hendryx]{gunjal2025rubrics}
Anisha Gunjal, Anthony Wang, Elaine Lau, Vaskar Nath, Yunzhong He, Bing Liu, and Sean Hendryx.
\newblock Rubrics as rewards: Reinforcement learning beyond verifiable domains.
\newblock \emph{arXiv preprint arXiv:2507.17746}, 2025.

\bibitem[Guo et~al.(2025)Guo, Yang, Zhang, Song, Wang, Zhu, Xu, Zhang, Ma, Bi, et~al.]{guo2025deepseek}
Daya Guo, Dejian Yang, Haowei Zhang, Junxiao Song, Peiyi Wang, Qihao Zhu, Runxin Xu, Ruoyu Zhang, Shirong Ma, Xiao Bi, et~al.
\newblock Deepseek-r1 incentivizes reasoning in llms through reinforcement learning.
\newblock \emph{Nature}, 645\penalty0 (8081):\penalty0 633--638, 2025.

\bibitem[Hou et~al.(2025)Hou, Hu, Li, Lu, Tang, and Dong]{hou2025treerl}
Zhenyu Hou, Ziniu Hu, Yujiang Li, Rui Lu, Jie Tang, and Yuxiao Dong.
\newblock Treerl: Llm reinforcement learning with on-policy tree search.
\newblock In \emph{Proceedings of the 63rd Annual Meeting of the Association for Computational Linguistics (Volume 1: Long Papers)}, pages 12355--12369, 2025.

\bibitem[Huang et~al.(2025)Huang, Zhuang, Lu, Qin, Xu, Zhao, Peng, Hu, Shen, Hu, Gu, Tu, Liu, Chen, Fu, Fan, Gu, Wang, Yang, Li, and Zhao]{huang2025reinforcementlearningrubricanchors}
Zenan Huang, Yihong Zhuang, Guoshan Lu, Zeyu Qin, Haokai Xu, Tianyu Zhao, Ru~Peng, Jiaqi Hu, Zhanming Shen, Xiaomeng Hu, Xijun Gu, Peiyi Tu, Jiaxin Liu, Wenyu Chen, Yuzhuo Fu, Zhiting Fan, Yanmei Gu, Yuanyuan Wang, Zhengkai Yang, Jianguo Li, and Junbo Zhao.
\newblock Reinforcement learning with rubric anchors, 2025.
\newblock URL \url{https://arxiv.org/abs/2508.12790}.

\bibitem[Jain et~al.(2025{\natexlab{a}})Jain, Gonzalez-Pumariega, Chen, Rush, Zhao, and Choudhury]{jain2025multi}
Arnav~Kumar Jain, Gonzalo Gonzalez-Pumariega, Wayne Chen, Alexander~M Rush, Wenting Zhao, and Sanjiban Choudhury.
\newblock Multi-turn code generation through single-step rewards.
\newblock \emph{arXiv preprint arXiv:2502.20380}, 2025{\natexlab{a}}.

\bibitem[Jain et~al.(2025{\natexlab{b}})Jain, Gu, Li, Yan, Zhang, Wang, Solar-Lezama, Sen, and Stoica]{jain2025livecodebench}
Naman Jain, Alex Gu, Wen-Ding Li, Fanjia Yan, Tianjun Zhang, Sida Wang, Armando Solar-Lezama, Koushik Sen, and Ion Stoica.
\newblock Livecodebench: Holistic and contamination free evaluation of large language models for code.
\newblock In \emph{International Conference on Learning Representations}, volume 2025, pages 58791--58831, 2025{\natexlab{b}}.

\bibitem[Jimenez et~al.(2024)Jimenez, Yang, Wettig, Yao, Pei, Press, and Narasimhan]{jimenez2024swe}
Carlos~E Jimenez, John Yang, Alexander Wettig, Shunyu Yao, Kexin Pei, Ofir Press, and Karthik Narasimhan.
\newblock Swe-bench: Can language models resolve real-world github issues?
\newblock In \emph{International Conference on Learning Representations}, volume 2024, pages 54107--54157, 2024.

\bibitem[Jin et~al.(2025)Jin, Xu, Li, Han, Zhou, Li, and Bai]{jin2025reveal}
Yiyang Jin, Kunzhao Xu, Hang Li, Xueting Han, Yanmin Zhou, Cheng Li, and Jing Bai.
\newblock Reveal: Self-evolving code agents via reliable self-verification.
\newblock \emph{arXiv preprint arXiv:2506.11442}, 2025.

\bibitem[Li et~al.(2025)Li, Sun, Wang, Li, and Shum]{li2025cuda}
Xiaoya Li, Xiaofei Sun, Albert Wang, Jiwei Li, and Chris Shum.
\newblock Cuda-l1: Improving cuda optimization via contrastive reinforcement learning.
\newblock \emph{arXiv preprint arXiv:2507.14111}, 2025.

\bibitem[Liu et~al.(2023)Liu, Xia, Wang, and Zhang]{liu2023your}
Jiawei Liu, Chunqiu~Steven Xia, Yuyao Wang, and Lingming Zhang.
\newblock Is your code generated by chatgpt really correct? rigorous evaluation of large language models for code generation.
\newblock \emph{Advances in neural information processing systems}, 36:\penalty0 21558--21572, 2023.

\bibitem[Liu et~al.(2026)Liu, Xu, Li, Zheng, Li, Liu, and He]{liu2026dr}
Wei Liu, Jiawei Xu, Yingru Li, Longtao Zheng, Tianjian Li, Qian Liu, and Junxian He.
\newblock Dr. kernel: Reinforcement learning done right for triton kernel generations.
\newblock \emph{arXiv preprint arXiv:2602.05885}, 2026.

\bibitem[Liu et~al.(2025)Liu, Chen, Li, Qi, Pang, Du, Lee, and Lin]{liu2025understanding}
Zichen Liu, Changyu Chen, Wenjun Li, Penghui Qi, Tianyu Pang, Chao Du, Wee~Sun Lee, and Min Lin.
\newblock Understanding r1-zero-like training: A critical perspective.
\newblock \emph{arXiv preprint arXiv:2503.20783}, 2025.

\bibitem[Lozhkov et~al.(2024)Lozhkov, Li, Allal, Cassano, Lamy-Poirier, Tazi, Tang, Pykhtar, Liu, Wei, et~al.]{lozhkov2024starcoder}
Anton Lozhkov, Raymond Li, Loubna~Ben Allal, Federico Cassano, Joel Lamy-Poirier, Nouamane Tazi, Ao~Tang, Dmytro Pykhtar, Jiawei Liu, Yuxiang Wei, et~al.
\newblock Starcoder 2 and the stack v2: The next generation.
\newblock \emph{arXiv preprint arXiv:2402.19173}, 2024.

\bibitem[Ouyang et~al.(2025)Ouyang, Guo, Arora, Zhang, Hu, R{\'e}, and Mirhoseini]{ouyang2025kernelbench}
Anne Ouyang, Simon Guo, Simran Arora, Alex~L Zhang, William Hu, Christopher R{\'e}, and Azalia Mirhoseini.
\newblock Kernelbench: Can llms write efficient gpu kernels?
\newblock \emph{arXiv preprint arXiv:2502.10517}, 2025.

\bibitem[Schulman et~al.(2017)Schulman, Wolski, Dhariwal, Radford, and Klimov]{schulman2017proximal}
John Schulman, Filip Wolski, Prafulla Dhariwal, Alec Radford, and Oleg Klimov.
\newblock Proximal policy optimization algorithms.
\newblock \emph{arXiv preprint arXiv:1707.06347}, 2017.

\bibitem[Sheng et~al.(2024)Sheng, Zhang, Ye, Wu, Zhang, Zhang, Peng, Lin, and Wu]{sheng2024hybridflow}
Guangming Sheng, Chi Zhang, Zilingfeng Ye, Xibin Wu, Wang Zhang, Ru~Zhang, Yanghua Peng, Haibin Lin, and Chuan Wu.
\newblock Hybridflow: A flexible and efficient rlhf framework.
\newblock \emph{arXiv preprint arXiv: 2409.19256}, 2024.

\bibitem[Team et~al.(2025)Team, Bai, Bao, Charles, Chen, Chen, Chen, Chen, Chen, Chen, et~al.]{team2025kimi}
Kimi Team, Yifan Bai, Yiping Bao, Y~Charles, Cheng Chen, Guanduo Chen, Haiting Chen, Huarong Chen, Jiahao Chen, Ningxin Chen, et~al.
\newblock Kimi k2: Open agentic intelligence.
\newblock \emph{arXiv preprint arXiv:2507.20534}, 2025.

\bibitem[Viswanathan et~al.(2026)Viswanathan, Sun, Kong, Cao, Neubig, and Wu]{viswanathan2026checklists}
Vijay Viswanathan, Yanchao Sun, Xiang Kong, Meng Cao, Graham Neubig, and Sherry Wu.
\newblock Checklists are better than reward models for aligning language models.
\newblock \emph{Advances in Neural Information Processing Systems}, 38:\penalty0 114728--114754, 2026.

\bibitem[Wei et~al.(2023)Wei, Wang, Liu, Ding, and Zhang]{wei2023magicoder}
Yuxiang Wei, Zhe Wang, Jiawei Liu, Yifeng Ding, and Lingming Zhang.
\newblock Magicoder: Empowering code generation with oss-instruct.
\newblock \emph{arXiv preprint arXiv:2312.02120}, 2023.

\bibitem[Xiao et~al.(2026)Xiao, Xia, Yang, Gao, Shen, Zhang, He, Lou, Luo, Wang, et~al.]{xiao2026mimo}
Bangjun Xiao, Bingquan Xia, Bo~Yang, Bofei Gao, Bowen Shen, Chen Zhang, Chenhong He, Chiheng Lou, Fuli Luo, Gang Wang, et~al.
\newblock Mimo-v2-flash technical report.
\newblock \emph{arXiv preprint arXiv:2601.02780}, 2026.

\bibitem[Xu et~al.(2025)Xu, Liu, Yin, Zhou, and Poovendran]{xu2025kodcode}
Zhangchen Xu, Yang Liu, Yueqin Yin, Mingyuan Zhou, and Radha Poovendran.
\newblock Kodcode: A diverse, challenging, and verifiable synthetic dataset for coding.
\newblock In \emph{Findings of the Association for Computational Linguistics: ACL 2025}, pages 6980--7008, 2025.

\bibitem[Yang et~al.(2025{\natexlab{a}})Yang, Li, Yang, Zhang, Hui, Zheng, Yu, Gao, Huang, Lv, et~al.]{yang2025qwen3}
An~Yang, Anfeng Li, Baosong Yang, Beichen Zhang, Binyuan Hui, Bo~Zheng, Bowen Yu, Chang Gao, Chengen Huang, Chenxu Lv, et~al.
\newblock Qwen3 technical report.
\newblock \emph{arXiv preprint arXiv:2505.09388}, 2025{\natexlab{a}}.

\bibitem[Yang et~al.(2025{\natexlab{b}})Yang, Guo, Huang, Liang, Wang, and Tang]{yang2025treerpo}
Zhicheng Yang, Zhijiang Guo, Yinya Huang, Xiaodan Liang, Yiwei Wang, and Jing Tang.
\newblock Treerpo: Tree relative policy optimization.
\newblock \emph{arXiv preprint arXiv:2506.05183}, 2025{\natexlab{b}}.

\bibitem[Yu et~al.(2026)Yu, Zhang, Zhu, Yuan, Zuo, Yue, Dai, Fan, Liu, Liu, et~al.]{yu2026dapo}
Qiying Yu, Zheng Zhang, Ruofei Zhu, Yufeng Yuan, Xiaochen Zuo, Yu~Yue, Weinan Dai, Tiantian Fan, Gaohong Liu, Lingjun Liu, et~al.
\newblock Dapo: An open-source llm reinforcement learning system at scale.
\newblock \emph{Advances in Neural Information Processing Systems}, 38:\penalty0 113222--113244, 2026.

\bibitem[Yue et~al.(2025)Yue, Yuan, Yu, Zuo, Zhu, Xu, Chen, Wang, Fan, Du, et~al.]{yue2504vapo}
Yu~Yue, Yufeng Yuan, Qiying Yu, Xiaochen Zuo, Ruofei Zhu, Wenyuan Xu, Jiaze Chen, Chengyi Wang, T~Fan, Z~Du, et~al.
\newblock Vapo: Efficient and reliable reinforcement learning for advanced reasoning tasks.
\newblock \emph{URL https://arxiv. org/abs/2504.05118}, 2025.

\bibitem[Zeng et~al.(2026)Zeng, Lv, Hou, Du, Zheng, Chen, Yin, Ge, Huang, Xie, et~al.]{zeng2026glm}
Aohan Zeng, Xin Lv, Zhenyu Hou, Zhengxiao Du, Qinkai Zheng, Bin Chen, Da~Yin, Chendi Ge, Chenghua Huang, Chengxing Xie, et~al.
\newblock Glm-5: from vibe coding to agentic engineering.
\newblock \emph{arXiv preprint arXiv:2602.15763}, 2026.

\bibitem[Zheng et~al.(2025)Zheng, Liu, Li, Chen, Yu, Gao, Dang, Liu, Men, Yang, et~al.]{zheng2025group}
Chujie Zheng, Shixuan Liu, Mingze Li, Xiong-Hui Chen, Bowen Yu, Chang Gao, Kai Dang, Yuqiong Liu, Rui Men, An~Yang, et~al.
\newblock Group sequence policy optimization.
\newblock \emph{arXiv preprint arXiv:2507.18071}, 2025.

\end{thebibliography}



\end{document}